\documentclass[letterpaper]{article} % DO NOT CHANGE THIS
\usepackage{aaai2026}  
\usepackage{times}  % DO NOT CHANGE THIS
\usepackage{helvet}  % DO NOT CHANGE THIS
\usepackage{courier}  % DO NOT CHANGE THIS
\usepackage[hyphens]{url}  % DO NOT CHANGE THIS
\usepackage{graphicx} % DO NOT CHANGE THIS
\usepackage{natbib}  % DO NOT CHANGE THIS AND DO NOT ADD ANY OPTIONS TO IT
\usepackage{caption} % DO NOT CHANGE THIS AND DO NOT ADD ANY OPTIONS TO IT
\usepackage{algorithm}
\usepackage{algorithmic}
\usepackage{appendix}
\usepackage{pdfpages}
\usepackage{newfloat}
\usepackage{listings}
\usepackage{amsmath}
\usepackage{booktabs}

\DeclareCaptionStyle{ruled}{labelfont=normalfont,labelsep=colon,strut=off} % DO NOT CHANGE THIS
\floatstyle{ruled}
\newfloat{listing}{tb}{lst}{}
\floatname{listing}{Listing}
\title{KG-Augmented Executable CoT for Mathematical Coding}

\author {
    Xingyu Chen\textsuperscript{\rm 1},
    Junxiu An\textsuperscript{\rm 1},
    Jun Guo\textsuperscript{\rm 2},
    Li Wang\textsuperscript{\rm 3}
    Jingcai Guo\textsuperscript{\rm 4}
}
\affiliations {
    \textsuperscript{\rm 1}School of Software Engineering, Chengdu University of Information Technology, Chengdu, China\\
    \textsuperscript{\rm 2}College of Applied Mathematics, Chengdu University of Information Technology, Chengdu, China\\
    \textsuperscript{\rm 3}School of Artificial Intelligence, Beihang University, Beijing, China\\
    \textsuperscript{\rm 4}Department of Computing, The Hong Kong Polytechnic University, Hong Kong, China\\

}
\begin{document}

\maketitle

\begin{abstract}
In recent years, large language models (LLMs) have excelled in natural language processing tasks but face significant challenges in complex reasoning tasks such as mathematical reasoning and code generation. To address these limitations, we propose KG-Augmented Executable Chain-of-Thought (KGA-ECoT), a novel framework that enhances code generation through knowledge graphs and improves mathematical reasoning via executable code. KGA-ECoT decomposes problems into a Structured Task Graph, leverages efficient GraphRAG for precise knowledge retrieval from mathematical libraries, and generates verifiable code to ensure computational accuracy. Evaluations on multiple mathematical reasoning benchmarks demonstrate that KGA-ECoT significantly outperforms existing prompting methods, achieving absolute accuracy improvements ranging from several to over ten percentage points. Further analysis confirms the critical roles of GraphRAG in enhancing code quality and external code execution in ensuring precision. These findings collectively establish KGA-ECoT as a robust and highly generalizable framework for complex mathematical reasoning tasks.
\end{abstract}


\section{Introduction}

Large Language Models (LLMs), such as ChatGPT\cite{achiam2023gpt} and Llama\cite{grattafiori2024llama}, have revolutionized natural language understanding and generation. They possess deep language comprehension, human-like text generation capabilities, contextual awareness, and robust problem-solving skills, making them invaluable in various domains\cite{yao2024survey}. However, it remains unclear whether LLMs reason based on deep logical understanding or rely on superficial pattern matching, limiting their performance in tasks requiring complex reasoning, such as mathematical problem-solving and code generation\cite{wang2023can}. These tasks demand precise logical reasoning and manipulation of domain-specific knowledge, such as mathematical theorems or programming libraries, which LLMs often struggle to handle accurately.

Mathematical reasoning and code generation present unique challenges for LLMs. First, mathematical problems often require multi-step reasoning, where errors in intermediate steps, such as misinterpreting units or algebraic expressions, can lead to incorrect solutions\cite{cobbe2021training}. Second, generating executable code demands a comprehensive understanding of programming syntax, library functionalities, and problem-specific constraints, which LLMs may fail to integrate without external guidance\cite{chen2021evaluating}. Third, LLMs are not consistently robust in pronoun fidelity. Even in simple, natural contexts, variations in sentence phrasing for the same meaning can lead to significant differences in the model's recognition accuracy\cite{gautam2024robust,miao2025autonomous}. These limitations underscore the need for structured frameworks to enhance LLMs’ ability to solve mathematical problems through accurate and verifiable code generation\cite{gao2023retrieval}.

To address the challenges in complex reasoning tasks, Chain-of-Thought (CoT) prompting has emerged as a powerful technique to elicit reasoning capabilities in LLMs. By guiding models through step-by-step reasoning, CoT significantly improves performance on logical and arithmetic tasks\cite{CoT,sprague2024cot,zhao2024expel}. Building on CoT, proposed Graph of Thoughts (GoT), which organizes reasoning steps into a graph structure, enabling more flexible and synergistic problem-solving\cite{GoT}. However, CoT and GoT primarily rely on textual reasoning, which lacks the precision and verifiability offered by executable code, especially for mathematical tasks requiring numerical computations or symbolic manipulations\cite{nye2021show,azerbayev2023llemma}.

Although CoT and GoT demonstrate remarkable performance in reasoning tasks, their reliance on textual reasoning limits their applicability in scenarios requiring external knowledge or precise computations. To address this, Retrieval-Augmented Generation (RAG) enhances the accuracy, timeliness, and reliability of generated content by retrieving relevant information from external knowledge bases \cite{lewis2020retrieval,gao2023retrieval}. However, traditional RAG struggles with global queries over large corpora due to its query-focused summarization nature, limiting its effectiveness in complex reasoning scenarios. To address this, GraphRAG leverages knowledge graphs to capture semantic relationships, enabling more precise and context-aware information retrieval\cite{edge2024local}. Despite its promise, GraphRAG remains underexplored in code generation tasks, where integrating domain-specific knowledge, such as mathematical libraries like SciPy\cite{virtanen2020scipy}, is critical for accurate and robust solutions.

To enhance the reasoning capabilities of LLMs in mathematical tasks, existing methods primarily rely on prompting techniques or task-specific fine-tuning\cite{brown2020language,ouyang2022training,qu2025latent}. However, these approaches often exhibit limited robustness and poor generalization across diverse tasks\cite{kirk2023understanding,hu2023llm,kumar2025llm}. To address these challenges, methods like Plan-and-Solve Prompting\cite{wang2023plan} guide LLMs to plan and solve mathematical problems step-by-step, improving reasoning performance without fine-tuning.However, such approaches rely on textual reasoning, lacking the precision and verifiability of executable code. To address these limitations, we propose a novel framework, KGA-ECoT, which integrates structured reasoning, knowledge-enhanced retrieval, and executable code generation. The framework leverages prompt-based strategies and agent-driven workflows, eliminating the need for fine-tuning while achieving significant performance gains on mathematical reasoning benchmarks.

\section{Related work}
Recent advancements in large language models (LLMs) have underscored their limitations in multi-step reasoning tasks, prompting researchers to explore structured planning frameworks. This section reviews two key techniques relevant to our approach: Chain-of-Thought (CoT)-based reasoning frameworks (e.g., CODEPLAN) and knowledge graph-augmented retrieval-augmented generation (GraphRAG).

\subsection{Structured Reasoning with Chain-of-Thought}
The core idea of CoT is to decompose a problem into multiple intermediate reasoning steps, generating a chain of reasoning in natural language to progressively derive the final answer. However, CoT-generated reasoning chains typically exist as free-form text, lacking structured representation, which can lead to logical leaps or errors in intermediate steps, particularly in mathematically intensive reasoning tasks\cite{zhou2022least}. To address CoT’s limitations, \cite{wen2024codeplan} proposed CODEPLAN, a scalable framework that enhances LLMs’ reasoning capabilities by generating and following code-based plans. By dividing the generation process into two stages—planning and surface realization—CODEPLAN produces a task-specific code-based plan outlining high-level reasoning steps. Subsequently, the model leverages the structured nature of code to capture complex semantics and control flows, performing internal reasoning and computations based on its generated pseudocode plan until the final answer is derived. This process is formally represented as:
\[ \text{CODEPLAN}(P) = \text{Plan}(P) \rightarrow \text{Execute}(\text{Plan}(P)) \]

where $P$ denotes the input problem, $\text{Plan}(P)$ generates the pseudocode plan, and $\text{Execute}(\cdot)$ performs reasoning based on the plan. Similarly, GoT models the reasoning process as a directed graph, using graph structures to represent reasoning steps, thereby enhancing CoT’s modularity. However, despite the success of CODEPLAN and GoT, their code-based plans remain non-executable pseudocode, fundamentally relying on LLMs’ “word-by-word” generation capabilities rather than true program execution. The authors also note that current LLMs face challenges in generating fully executable code plans, as this requires comprehensive knowledge of libraries, APIs, and domain-specific contexts. This limitation highlights the gap between structured planning and actual code functionality. The shortcomings of these approaches suggest that relying solely on LLMs’ text or graph generation capabilities is insufficient for achieving precision and executability in complex reasoning tasks, motivating our design of the KGA-ECoT framework, which integrates knowledge graphs and executable code.

\subsection{Graph-based Retrieval-Augmented Generation}
 Graph-based Retrieval-Augmented Generation(GraphRAG) is a cutting-edge technique that integrates knowledge graphs\cite{peng2023knowledge} with retrieval-augmented generation (RAG) to enhance matching degree between the retrieved content and the questions. \cite{gautam2024robust} and \cite{peng2024graph} note that, compared to traditional RAG methods, GraphRAG more effectively captures semantic relationships and contextual dependencies. GraphRAG constructs a knowledge graph $ G = (V, E) $, where $ V $ represents entity nodes and $ E $ denotes relational edges, by extracting entities and relationships from text. It then employs graph neural networks (GNNs, \cite{wu2022graph}) to aggregate node information and generate node embeddings:
\[ h_v = \text{GNN}(G, v), \quad v \in V \]

Compared to traditional RAG, which relies on vector embeddings, GraphRAG employs the Leiden hierarchical algorithm \cite{traag2019louvain} and graph traversal techniques (e.g., BFS \cite{bundy1984breadth} or PageRank variants \cite{page1999pagerank}) for indexing, thereby achieving more precise knowledge retrieval.It excels particularly in long-document understanding and multi-hop reasoning tasks, demonstrating higher accuracy and interpretability, especially in scenarios requiring cross-document information integration.

However, \cite{gautam2024robust} highlight that GraphRAG’s retrieval precision may be constrained by the heterogeneity between node embeddings and query embeddings. Node embeddings in knowledge graphs depend on the graph’s topological structure (e.g., neighboring nodes and edge weights), excelling at capturing structured relationships among entities but potentially lacking in semantic expressiveness. In contrast, query embeddings, generated by pretrained language models (e.g., BERT \cite{devlin2019bert} or its variants), rely heavily on textual semantics and context, excelling at capturing semantic similarities in words and sentences but struggling to directly reflect the graph’s structured information. Due to their differing generation mechanisms, node and query embeddings may exhibit inconsistent distributions in feature space\cite{jiang2020co}, leading to reduced retrieval precision due to semantic and structural heterogeneity:
\[ \text{Loss}_{\text{mismatch}} = \text{Dist}(E_{\text{node}}, E_{\text{query}}) \]

where $ E_{\text{node}} $ and $ E_{\text{query}} $ represent node and query embeddings, respectively, and $ \text{Dist}(\cdot) $ measures the distributional disparity between them.

\subsection{Contributions}
Frameworks like CODEPLAN and GoT enhance reasoning capabilities through structured representations, but their reasoning still relies on the text generation of LLMs, which limits the precise computing power. GraphRAG enhances the retrieval function through knowledge graphs, but the inconsistency of the embedding space hinders code generation, causing a disconnection between structured reasoning and retrieval. To address these shortcomings and fully leverage their respective advantages, we propose KGA-ECoT, which combines the structured reasoning of CoT with the knowledge-enhanced retrieval of GraphRAG, while integrating executable code generation. This approach overcomes their individual limitations. By improving the accuracy of node query embeddings through efficient semantic alignment, KGA-ECoT seamlessly integrates task decomposition, knowledge retrieval, and code execution, achieving significant performance improvements in mathematical reasoning benchmark tests.

The main contributions of this paper are as follows:
\begin{itemize}
\item KGA-ECoT Framework: We propose a structured CoT approach that integrates knowledge graphs and executable code generation. By constructing task graphs and generating executable code, KGA-ECoT overcomes the limitations of LLMs’reliance on sequential text-based reasoning, significantly improving computational capability and solution accuracy in complex reasoning tasks.

\item Innovative Hierarchical Graph Embedding: To address the matching precision issue caused by the heterogeneity between node and query embeddings, we introduce a graph embedding technique that fuses textual embeddings with graph structural information, substantially enhancing GraphRAG’s retrieval efficiency and code generation quality.

\item Empirical Validation: We demonstrate the superior performance of KGA-ECoT across multiple mathematical reasoning benchmark datasets and various scales and types of base models, highlighting its potential in complex reasoning tasks.
\end{itemize}

\section{Method}
\subsection{Problem definition}

\begin{figure*}[t] % 使用 figure* 环境
\centering
\includegraphics[width=0.9\textwidth]{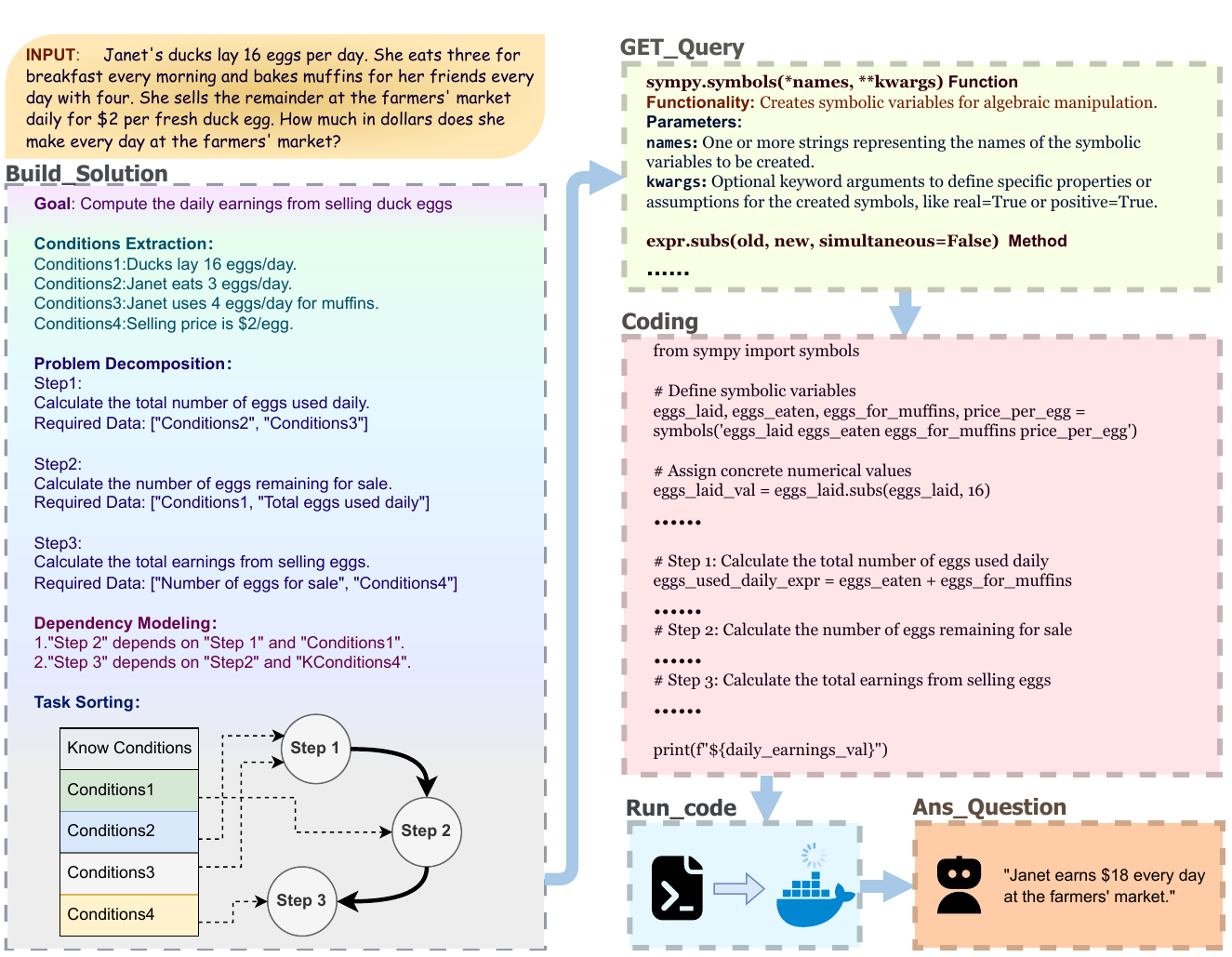}  % Reduce the figure size so that it is slightly narrower than the column.
\caption{CoT for Mathematical Coding(This figure demonstrates a structured approach to solving mathematical word problems, using the example of calculating daily revenue from selling duck eggs. The framework includes the following main stages: (1) INPUT: Input the mathematical problem. (2) Build\textunderscore Solution: This section details the process of problem decomposition and solution path construction, including: Goal, Conditions Extraction, Problem Decomposition, Dependency Modeling, and Task Sorting. The GET\textunderscore Query section retrieves tools and functions for generating relevant code from a knowledge graph. (3) Coding: This section converts the problem-solving steps into executable code. (4) Run\textunderscore code: Represents the execution stage of the generated code. (5) Ans\textunderscore Question: This section polishes and outputs the final answer obtained from execution.)}
\label{fig1}
\end{figure*}

Given a mathematical problem $Q$, the goal is to generate executable code $C$, which, when executed, produces the final answer $A$. We model the problem-solving process as a task graph $G$, where each node represents a reasoning subtask, and transitions between states are determined by task dependencies, guiding code generation and reasoning execution.

\subsection{CoT for Mathematical Coding}
The KGA-ECoT framework aims to enhance the capability of LLMs in mathematical reasoning tasks by generating executable code, overcoming the limitations of traditional CoT that rely on sequential text-based reasoning. As shown in Figure 1, this framework models the problem-solving process as a structured CoT pipeline, comprising five key nodes: Build\textunderscore Solution, GET\textunderscore Query, Coding, Run\textunderscore code, and Ans\textunderscore Question---to systematically decompose problems, generate code, and validate answers. By integrating a GraphRAG-indexed knowledge base, KGA-ECoT significantly improves the quality of code generation.

The Build\textunderscore Solution node is responsible for decomposing the mathematical problem into structured subtasks and generating a task graph to guide subsequent reasoning. This process includes five steps: (1) Goal Analysis: Identify the main objective and problem type (e.g., computing a value or solving for a variable), listing key elements. (2)Conditions Extraction: Record given values or constraints, including all known information, numerical values, constants, and any explicit assumptions or restrictions in the problem statement, serving as the basis for subsequent computations and reasoning. (3) Problem Decomposition: Divide the problem into subtasks. For example, in Figure 1, “calculate the total number of eggs used daily” and “calculate the number of eggs available for sale.”(4) Dependency Modeling: Analyze dependencies among subtasks to determine the execution order. (5) Task Sorting: Generate a task sequence based on dependencies, forming a task graph where nodes represent subtasks and edges denote dependency relationships.

The GET\textunderscore Query node leverages a hierarchical graph embedding method (see Section 3.3) to retrieve the function descriptions and other relevant information needed for generating code from the knowledge base, in order to support high-quality code generation. The Coding node generates executable Python code based on the structured task graph and knowledge retrieved by GET\textunderscore Query. The Run\textunderscore code node executes generated Python code within a Docker-based isolated runtime environment, ensuring secure and stable execution. Leveraging Docker tools, it automatically runs the code and captures the results, which are then passed as intermediate outputs to subsequent nodes for further processing. If code execution fails (e.g., due to syntax errors or runtime exceptions), Run\textunderscore code logs the error information for further processing by the Ans\textunderscore Question node. The Ans\textunderscore Question node serves as the final safeguard in the CoT pipeline, verifying the consistency of the code output with the problem requirements. In cases of code failure, it generates an answer using existing reasoning information to ensure robustness. Compared to traditional CoT, KGA-ECoT enhances reasoning precision through executable code and validation mechanisms.

\subsection{Hierarchical graph embedding}
The code generated by large language models often encounters some issues, causing it to fail to execute. The reasons may include changes in function versions, etc. SymPy \cite{meurer2017sympy} is a commonly used mathematical reasoning library. In this paper, we built a directed acyclic knowledge graph based on the official SymPy documentation, which includes 268 directory nodes and 3923 callable nodes, to support knowledge retrieval for mathematical programming tasks.To systematically organize SymPy’s functionality, we represent its structure as a bipartite graph comprising Catalogue Nodes and Callable Nodes, as shown in Figure 2. Catalogue Nodes form a hierarchical framework, structuring functional modules and documentation entries into a multi-layered directory while providing concise summaries and descriptions of the methods within each layer. Callable Nodes encapsulate specific implementations (functions, classes, or methods) along with detailed parameter and return specifications. By grouping Callable Nodes under multi-layered Catalogue Nodes, the framework enables precise contextual reasoning. Callable Nodes within the same leaf-level directory exhibit strong functional coherence, while the hierarchical structure effectively disambiguates homonymous functions across categories, thereby enhancing the language model’s accuracy in knowledge retrieval and reasoning.

Traditional GraphRAG suffers from limited retrieval precision due to the semantic heterogeneity between node embeddings and query embeddings. Leveraging the directed acyclic nature and hierarchical organization of the knowledge graph, where parent nodes summarize and abstract their child nodes, we propose a hierarchical graph embedding method. This approach generates node embeddings with both semantic consistency and topology awareness through tree-based feature propagation and weighted feature fusion, enhancing knowledge retrieval efficiency in the GET\textunderscore Query stage. Compared to traditional GraphRAG’s query parsing and node embedding methods, our approach exploits the graph’s hierarchical structure, simplifying the embedding generation process and improving retrieval precision.

In the node embedding generation phase, we propose a hierarchical feature propagation algorithm to integrate initial semantic embeddings generated by the bge-m3 model with graph structural information, producing optimized node embeddings (Graph\textunderscore Embd). As shown in Figure 2, this algorithm fully leverages the hierarchical topology of Catalogue Nodes and Callable Nodes in the graph. Specifically, the algorithm first generates initial embeddings based on node content (e.g., function declarations, functional descriptions) using bge-m3, followed by feature propagation in the hierarchical order depicted in Figure 2 (from root to leaf nodes). For non-root nodes, the final embedding $e_n$ is computed as:
\[ e_n = w \cdot x_n + (1 - w) \cdot e_p \]

where $ x_n $ is the node’s initial bge-m3 semantic embedding, $ e_p $ is the parent node’s embedding, and $ w \in [0,1] $ is an adjustable weighting coefficient. By tuning $ w $, we can flexibly control the balance between the node’s local semantic features and the parent node’s contextual information: a smaller $ w $ emphasizes the node’s local semantic characteristics, while a larger $ w $ better incorporates the global contextual information inherent in the Catalogue Node. This hierarchical feature fusion mechanism ensures that the generated embeddings possess both semantic consistency and topology awareness.

\begin{figure}[htbp] % 或 [t], [h], [b], [p]
\centering
\includegraphics[width=0.95\columnwidth]{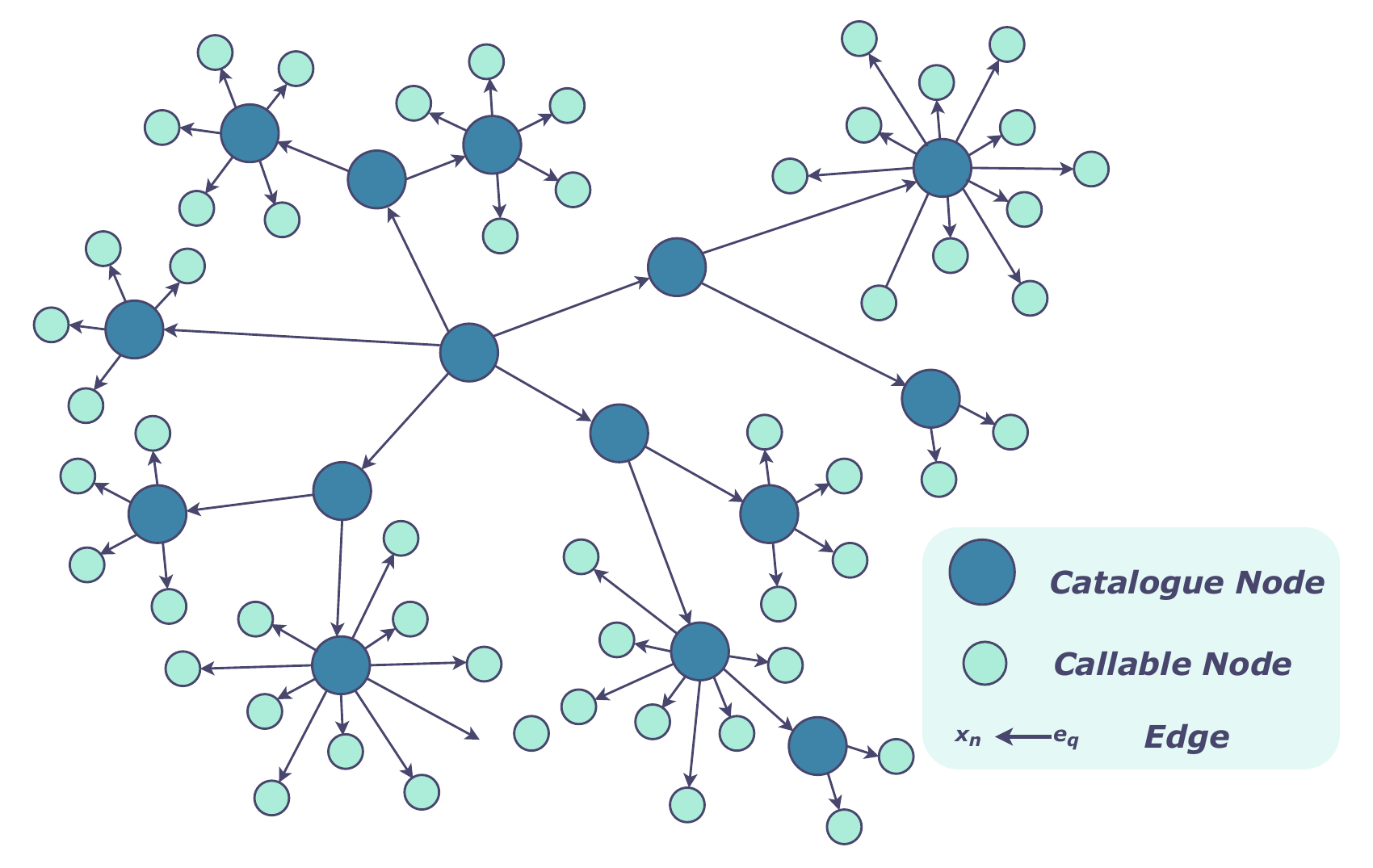} 
\caption{Hierarchical graph embedding}
\label{Figure/Hierarchical_embedding.pdf} % 给图片一个唯一的标签，方便在文中引用
\end{figure}

In the GET\textunderscore Query phase, query embeddings $e_q$ are encoded by bge-m3 based on the descriptions of subtasks in the structured task graph. Since the embeddings of Callable nodes have incorporated the semantic information of Catalogue nodes through hierarchical propagation, the system only needs to compute the cosine similarity between $ e_q $ and the Callable node embeddings (Graph\textunderscore Embd):
\[ \text{sim}(e_q, e_n) = \frac{e_q \cdot e_n}{\|e_q\| \cdot \|e_n\|} \]

The system then selects the top-$ n $ nodes with the highest similarity scores as retrieval results, which are passed to the code generation phase (Coding). This retrieval strategy fully exploits the semantic and topology-aware capabilities of Graph\textunderscore Embd, ensuring that the generated code is of high quality and precisely aligns with the actual task requirements.

Compared to the embedding schemes used in traditional GraphRAG, our approach significantly enhances the semantic consistency and topology awareness of node embeddings through hierarchical feature propagation and adaptive weighted fusion in directed acyclic graphs. By leveraging the properties of knowledge graphs, the embedding matching process is simplified, enabling the constructed knowledge graph to provide richer structured contextual information and precise functional associations for the language model. This effectively addresses the semantic ambiguity inherent in traditional methods, thereby enhancing the functional accuracy and practical applicability of the generated code.

\section{Experiment}
\subsection{Benchmarks}
The proposed method was evaluated on three distinct datasets: (1) GSM8K \cite{cobbe2021training}: Comprising 1,319 elementary math word problems, this dataset assesses multi-step arithmetic and logical reasoning in varied narrative contexts, requiring precise numerical inference. (2) MATH-500 \cite{lightman2023let}:A challenging set of 500 problems spanning algebra, geometry, number theory, and precalculus, testing advanced mathematical insight and complex theorem application. (3) SVAMP \cite{patel2021nlp}: Containing 300 problems designed to evaluate robustness against superficial pattern recognition, this dataset tests the ability to distinguish mathematically distinct problems and avoid shortcut learning.

To evaluate KGA-ECoT’s performance on mathematical reasoning and code generation tasks, we adopted tailored evaluation metrics for different datasets. For the GSM8K and SVAMP datasets, we used Exact Match Accuracy (EMA), which measures the proportion of model outputs that exactly match the reference answers. This metric is suitable for these datasets’ numerical or simple fractional answers \cite{rajpurkar2016squad}. For the MATH-500 dataset, since answers are output in LaTeX{} format and the same mathematical expression may have multiple representations,e.g., “\$\textbackslash dfrac\{a\}\{b\}\$” and “\$a/b\$”, we employed Semantic Equivalence Accuracy(SEA), where an answer is deemed correct if it is mathematically equivalent to the reference \cite{chen2021evaluating}.
\subsection{Main Result}
We evaluated our method on large language model backbones of varying scales: DeepSeek-7B (DS-7B), DeepSeek-14B (DS-14B), LLaMA3.2-3B (L3.2-3B), and LLaMA3.1-8B (L3.1-8B). These were compared against advanced prompt engineering baselines: Zero-shot Chain-of-Thought (CoT (0-shot)), Few-shot Chain-of-Thought (CoT (3-shot), with three examples), Prompt Selection (PS), and CODEPLAN (CP). All baselines used prompt engineering without fine-tuning the foundation models to ensure fair comparisons.

As shown in Tables 1 to 3, our method demonstrates consistent and significant performance improvements across all backbone models and most benchmark tests. This advantage is primarily attributed to KGA-ECoT’s innovations in reshaping the chain-of-thought, constructing task graphs, generating code, and leveraging knowledge-augmented GraphRAG mechanisms. Specifically, compared to traditional CoT and PS methods, which rely on the internal reasoning capabilities of LLMs, KGA-ECoT significantly enhances reasoning reliability and accuracy by transforming complex mathematical reasoning into solution steps and externally executable code, thereby achieving stable performance gains. More notably, KGA-ECoT outperforms nearly all advanced prompting methods, achieving an average performance improvement of 2.36 points on GSM8K, 3.59 points on MATH-500, and 2.37 points on SVAMP.

\begin{table}[h]
\centering
\setlength{\tabcolsep}{6pt} % 压缩列间距以适应版面
\begin{tabular}{lcccc}
\toprule
Method & DS-7B & DS-14B & L3.2-3B & L3.1-8B \\
\midrule
CoT (0) & 78.77\% & 83.78\% & 48.07\% & 57.92\% \\
CoT (3) & 83.70\% & 92.72\% & 52.54\% & \textbf{76.91\%} \\
PS      & 85.14\% & 91.51\% & 30.25\% & 57.62\% \\
CP      & 85.44\% & 91.43\% & 55.64\% & 56.71\% \\
Ours    & \textbf{87.49\%} & \textbf{92.80\%} & \textbf{63.91\%} & 75.97\% \\
\bottomrule
\end{tabular}
\caption{Model comparison on GSM8K dataset(EMA)} % 标题置于下方，10 点 Roman 字体
\label{tab:model_comparison_gsm8k}
\end{table}

\begin{table}[h]
\centering
\setlength{\tabcolsep}{6pt}
\begin{tabular}{lcccc}
\toprule
Method & DS-7B & DS-14B & L3.2-3B & L3.1-8B \\
\midrule
CoT (0) & 76.40\% & 86.60\% & 36.00\% & 39.40\% \\
CoT (3) & 70.60\% & 87.12\% & 37.02\% & 24.80\% \\
PS      & 79.80\% & 84.60\% & 40.40\% & 41.40\% \\
CP      & 82.60\% & 85.20\% & 26.20\% & 27.40\% \\
Ours    & \textbf{89.40\%} & \textbf{88.80\%} & \textbf{41.10\%} & \textbf{46.60\%} \\
\bottomrule
\end{tabular}
\caption{Model comparison on MATH-500 dataset(SEA)}
\label{tab:model_comparison_math500}
\end{table}

\begin{table}[h]
\centering
\setlength{\tabcolsep}{6pt}
\begin{tabular}{lcccc}
\toprule
Method & DS-7B & DS-14B & L3.2-3B & L3.1-8B \\
\midrule
CoT (0) & 87.60\% & 89.33\% & 70.67\% & 67.67\% \\
CoT (3) & 86.00\% & 91.67\% & 73.67\% & 82.67\% \\
PS      & 86.33\% & 91.00\% & 47.00\% & 66.70\% \\
CP      & 85.30\% & 90.67\% & 72.00\% & 72.67\% \\
Ours    & \textbf{89.00\%} & \textbf{92.33\%} & \textbf{79.09\%} & \textbf{84.67\%} \\
\bottomrule
\end{tabular}
\caption{Model comparison on SVAMP dataset(EMA)}
\label{tab:model_comparison_svamp}
\end{table}

This superiority is mainly due to KGA-ECoT’s integration of GraphRAG technology, which retrieves syntax libraries relevant to Python mathematical computations, effectively addressing CODEPLAN’s challenges with low code executability and the integration of domain-specific mathematical knowledge. This knowledge-augmented code generation and execution paradigm enables KGA-ECoT to produce higher-quality, more robust code to tackle a wide range of problems, from elementary arithmetic to advanced mathematical challenges, while effectively resisting interference from superficial patterns. These findings collectively validate the generality, effectiveness, and robustness of KGA-ECoT in enhancing the mathematical reasoning capabilities of large language models, providing solid experimental support for developing more powerful and reliable mathematical reasoning models.

\subsection{Ablation experiment}
To systematically evaluate the contributions of each core module in the KGA-ECoT method, we designed and conducted a series of ablation experiments. This section provides a detailed analysis of the impact on model performance when removing the GraphRAG module (KGA-ECoT without RAG) and the external code execution module (KGA-ECoT without coding), where KGA-ECoT without coding further removes the external code execution module on top of removing GraphRAG.

\begin{table}[h]
\centering
\setlength{\tabcolsep}{6pt}
\begin{tabular}{lcccc}
\toprule
Method & DS-7B & DS-14B & L3.2-3B & L3.1-8B \\
\midrule
KGA-ECoT     & \textbf{87.49\%} & 92.80\% & \textbf{63.91\%} & 75.97\% \\
Drop RAG     & 86.13\% & \textbf{93.25\%} & 62.70\% & \textbf{76.30\%} \\
Drop coding  & 79.91\% & 93.10\% & 57.39\% & 69.90\% \\
\bottomrule
\end{tabular}
\caption{Ablation comparison on GSM8K(EMA)}
\label{tab:ablation_gsm8k}
\end{table}

\begin{table}[h]
\centering
\setlength{\tabcolsep}{6pt}
\begin{tabular}{lcccc}
\toprule
Method & DS-7B & DS-14B & L3.2-3B & L3.1-8B \\
\midrule
KGA-ECoT & \textbf{89.40\%} & 88.80\% & 41.10\% & \textbf{46.60\%} \\
Drop RAG    & 85.30\% & \textbf{91.60\%} & \textbf{43.40\%} & 42.20\% \\
Drop coding & 84.80\% & 90.20\% & 32.60\% & 35.10\% \\
\bottomrule
\end{tabular}
\caption{Ablation comparison on MATH-500(SEA)}
\label{tab:ablation_math500}
\end{table}

\begin{table}[h]
\centering
\setlength{\tabcolsep}{6pt}
\begin{tabular}{lcccc}
\toprule
Method & DS-7B & DS-14B & L3.2-3B & L3.1-8B \\
\midrule
KGA-ECoT    & \textbf{89.33\%} & 92.33\% & \textbf{79.09\%} & \textbf{84.67\%} \\
Drop RAG    & 89.00\% & \textbf{94.67\%} & 76.00\% & 82.60\% \\
Drop coding & 80.30\% & 94.00\% & 73.67\% & 78.33\% \\
\bottomrule
\end{tabular}
\caption{Ablation comparison on SVAMP(EMA)}
\label{tab:ablation_svamp}
\end{table}

As shown in Tables 4 to 6, by comparing the performance of the full KGA-ECoT model with KGA-ECoT without RAG, we can clearly quantify the benefits of the GraphRAG module: removing GraphRAG generally leads to performance degradation, with its contribution being particularly pronounced on the MATH-500 dataset, which requires complex mathematical knowledge. This indicates that GraphRAG, by precisely retrieving Python mathematical computation-related syntax libraries, effectively enhances the quality, correctness, and executability of code generated by LLMs, especially when addressing problems involving advanced mathematical concepts, where it provides critical, domain-specific knowledge support. Although in certain cases (e.g., DeepSeek-14B on GSM8K and SVAMP), removing RAG results in slight performance improvements, this may be due to room for fine-tuning in the knowledge graph construction for specific models or samples, or the introduction of minor noise. However, the overall trend and significant gains on MATH-500 strongly demonstrate that GraphRAG is an indispensable knowledge augmentation module in KGA-ECoT.

By comparing KGA-ECoT without RAG with KGA-ECoT without coding, we can quantify the value of the external code execution module: removing the external code execution module (i.e., degrading from KGA-ECoT without RAG to KGA-ECoT without coding) results in the most significant performance drop. Across all models and datasets, accuracy consistently declines. For instance, on the DeepSeek-7B model, GSM8K accuracy plummets from 86.13\% to 79.91\%, MATH-500 drops from 85.30\% to 84.80\%, and SVAMP falls sharply from 89.00\% to 80.30\%. This decline is even more pronounced on the Llama 3 series, such as Llama3.2-3B on MATH-500, where accuracy drops from 43.40\% to 32.60\%, a decrease of 10.8 percentage points. This strongly proves that executing LLM-generated code via an external executor, rather than relying solely on the LLM’s internal text-based reasoning, is critical to ensuring precise mathematical problem-solving and verifiable answers. Even the most optimized prompt designs cannot substitute for the precision and determinacy of external computation.

In summary, the ablation experiment results clearly confirm that KGA-ECoT’s superior performance stems from the synergistic interaction of its components: carefully designed prompts lay a robust structural foundation for the entire reasoning process; external code execution provides unmatched computational precision and answer verification, serving as the fundamental guarantee for performance improvements; and GraphRAG significantly enhances the quality and domain adaptability of generated code through knowledge augmentation, particularly excelling in advanced mathematical problems. These findings not only validate the importance of each module in KGA-ECoT but also provide insights for future research and development of LLMs in the field of mathematical reasoning.

\section{Conclusion}
In this work, we introduce a novel Chain-of-Thought framework, KGA-ECoT, designed to significantly enhance the mathematical reasoning capabilities of LLMs. KGA-ECoT reformulates and partitions mathematical problems into a sequence of restructured subtasks through task graph construction, markedly improving problem clarity and logical coherence. Furthermore, it innovatively integrates GraphRAG technology to retrieve computation-related syntax libraries, thereby enhancing the model’s ability to generate solution code. Most importantly, KGA-ECoT’s design offers high flexibility and generality: it can tailor knowledge graphs to specific reasoning scenarios without requiring secondary fine-tuning of LLMs on large, broadly scoped datasets, thus greatly expanding its application scenarios and reducing deployment costs. Extensive evaluations across three commonly used mathematical reasoning datasets and various scales and types of backbone models demonstrate KGA-ECoT’s significant performance advantages, outperforming other advanced prompt engineering and code-based reasoning methods in the vast majority of tasks. Ablation analyses further confirm the critical contributions of the GraphRAG module in improving code executability and domain knowledge integration, as well as the indispensable role of the external code execution mechanism in ensuring computational precision and answer verifiability. These findings collectively position KGA-ECoT as a powerful, efficient, and highly generalizable mathematical reasoning framework, laying a solid foundation for the future development of more intelligent and reliable LLM-based mathematical problem-solving systems.

\bibliography{aaai2026}
\end{document}